\documentclass[sigconf]{acmart}

\AtBeginDocument{%
  \providecommand\BibTeX{{%
    \normalfont B\kern-0.5em{\scshape i\kern-0.25em b}\kern-0.8em\TeX}}}

\setcopyright{acmlicensed}
\copyrightyear{2024}
\acmYear{2024}

\acmConference[KDD '24] {Proceedings of the 30th ACM SIGKDD Conference on Knowledge Discovery and Data Mining }{August 25--29, 2024}{Barcelona, Spain.}
\acmBooktitle{Proceedings of the 30th ACM SIGKDD Conference on Knowledge Discovery and Data Mining (KDD '24), August 25--29, 2024, Barcelona, Spain}
\acmISBN{979-8-4007-0490-1/24/08}
\acmDOI{10.1145/3637528.3671616}
\usepackage{booktabs}
\usepackage{arydshln}
\usepackage{enumitem}
\usepackage{array}
\usepackage{algorithm}
\usepackage{algpseudocode}
\usepackage{amsmath}
\usepackage{subfigure}
\usepackage{graphicx}
\usepackage{multirow}
\usepackage[utf8]{inputenc} 
\usepackage{hyperref}       
\usepackage{url}            
\usepackage{booktabs}       
\usepackage{amsfonts}       
\usepackage{nicefrac}       
\usepackage{microtype}      
\usepackage{xcolor}         
\usepackage{balance}

\begin{document}

\title{Addressing Shortcomings in Fair Graph Learning Datasets: Towards a New Benchmark}

\author{Xiaowei Qian}
\authornote{These authors contributed equally to this work.}
\affiliation{
    \institution{Rensselaer Polytechnic Institute}
    \city{}
    \country{}
}
\email{xiaoweiqian0311@gmail.com}

\author{Zhimeng Guo}
\authornotemark[1]
\affiliation{
    \institution{The Pennsylvania State University}
    \city{}
    \country{}
}
\email{zhimeng@psu.edu}

\author{Jialiang Li}
\affiliation{
    \institution{New Jersey Institute of Technology}
    \city{}
    \country{}
}
\email{jl2356@njit.edu}

\author{Haitao Mao}
\affiliation{
    \institution{Michigan State University}
    \city{}
    \country{}
}
\email{haitaoma@msu.edu}

\author{Bingheng Li}
\affiliation{
    \institution{Michigan State University}
    \city{}
    \country{}
}
\email{libinghe@msu.edu}

\author{Suhang Wang}
\affiliation{
    \institution{The Pennsylvania State University}
    \city{}
    \country{}
}
\email{szw494@psu.edu}

\author{Yao Ma}
\authornote{Corresponding author.}
\affiliation{
    \institution{Rensselaer Polytechnic Institute}
    \city{}
    \country{}
}
\email{may13@rpi.edu}

\renewcommand{\shortauthors}{Qian and Guo, et al.}

\begin{abstract}
  
Fair graph learning plays a pivotal role in numerous practical applications. Recently, many fair graph learning methods have been proposed; however, their evaluation often relies on poorly constructed semi-synthetic datasets or substandard real-world datasets. In such cases, even a basic Multilayer Perceptron (MLP) can outperform Graph Neural Networks (GNNs) in both utility and fairness. In this work, we illustrate that many datasets fail to provide meaningful information in the edges, which may challenge the necessity of using graph structures in these problems. 
To address these issues, we develop and introduce a collection of synthetic, semi-synthetic, and real-world datasets that fulfill a broad spectrum of requirements. These datasets are thoughtfully designed to include relevant graph structures and bias information crucial for the fair evaluation of models. The proposed synthetic and semi-synthetic datasets offer the flexibility to create data with controllable bias parameters, thereby enabling the generation of desired datasets with user-defined bias values with ease. Moreover, we conduct systematic evaluations of these proposed datasets and establish a unified evaluation approach for fair graph learning models. Our extensive experimental results with fair graph learning methods across our datasets demonstrate their effectiveness in benchmarking the performance of these methods. Our datasets and the code for reproducing our experiments are available \footnote{\url{https://github.com/XweiQ/Benchmark-GraphFairness}}.

\end{abstract}




\maketitle

\section{Introduction}

Graph structure is ubiquitous language to model complicated relationships. As much information is organized as graph structure, graph neural networks are becoming increasingly important in various fields, including knowledge graphs~\cite{arora2020survey, zhang2020relational}, drug discovery~\cite{lin2020kgnn, xiong2021graph} and social media mining~\cite{fan2019graph}. GNNs are versatile in handling tasks related to graphs, enhancing performance in activities from node classification~\cite{wang2021mixup, zhang2023contrastive, kipf2016semi} to link prediction~\cite{shiao2023link, yunneognns} and graph classification~\cite{wang2022taskadaptive}. 
However, accompanying the wide deployment in many critical systems~\cite{feldman2015certifying}, concerns about the potential risks associated with GNNs are growing. Research shows that GNNs can either inherit or exacerbate bias in the data, leading to unfair and biased predictions, which potentially reinforce existing prejudices and discrimination~\cite{daicomprehensive, dai2021say}. This issue has raised ethical and societal concerns, significantly hindering GNNs' application in sensitive decision-making areas, such as ranking job applicants~\cite{mehrabi2021survey} and predicting criminal behavior~\cite{suresh2019framework}.  

To tackle the fairness challenge, a series of fair graph learning models have been developed, e.g., FairGNN~\cite{dai2021say}, NIFTY~\cite{agarwal2021unified}, and EDITS~\cite{dong2022edits}. These methods aim to improve fairness while maintaining the model's accuracy. Building on the foundational concerns regarding the development of fair graph learning models, it is crucial to scrutinize the existing evaluation frameworks that assess these models. Upon examination, we find that existing evaluation protocols suffer from several pitfalls that impede our ability to properly evaluate these methods, which are summarized as follows:
\begin{itemize}[leftmargin=*]
    \item The evaluation of fair graph learning models is often limited to a few poorly constructed semi-synthetic datasets~\cite{agarwal2021unified} converted from tabular datasets and an array of real-world datasets~\cite{dai2021say}. Specifically, the graph connections in the semi-synthetic datasets are constructed based on feature similarity, which often struggles to provide additional information for prediction. Our exploration in Section~\ref{sec:semi} (see results in Table~\ref{tab: old_semi_results}) demonstrates that, on the semi-synthetic datasets, \emph{Multilayer Perceptrons (MLPs), which do not leverage the graph structure outperforms GCN and fairness-focused GNN methods including NIFTY~\cite{agarwal2021unified} and FairGNN~\cite{dai2021say} by a decent margin}. In addition, not only do MLPs achieve higher performance, but they also maintain superior fairness metrics on these semi-synthetic datasets. This phenomenon is not limited to semi-synthetic datasets. Certain real-world datasets, which we discuss further in Table~\ref{tab: old_real_results}, exhibit similar patterns.  Furthermore, on some certain real-world datasets, the incorporation of graph structure does not introduce additional biases, as evidenced by GCNs and MLPs achieving comparable levels of fairness. These findings underline a critical issue with the current datasets used for evaluating fair graph learning models, i.e., the graph structure in these datasets does not effectively bring additional information. They do not provide the necessary challenges or contexts where the sophisticated mechanisms of fair GNNs can demonstrate their full potential and value over more basic analytical models.
    \item Unlike traditional node classification tasks where models cease training upon plateauing performance on a validation set, fair graph learning models must also consider fairness. Therefore, different model selection strategies have been developed to achieve this goal, which introduces additional complexities for evaluation. Specifically, our investigation in Section~\ref{sec: earlystopping} highlights that the choice of model selection strategy significantly affects both model performance and fairness. \emph{The diversity in model selection strategies makes it challenging to determine if observed differences in outcomes are the result of the models' inherent algorithmic qualities or the particular model selection mechanisms employed, complicating the process of accurately evaluating these models.} 
\end{itemize}

To address these concerns, we first propose to unify the model selection mechanisms used across various models. Following this, we aim to develop a collection of new datasets specifically designed for evaluating fair graph learning. In creating these datasets, we focus on two crucial aspects: (1) the Utility of Graph Structure, and (2) the Potential for Bias Amplification via Graph Structure. Our approach ensures that only models adept at leveraging graph structure while neutralizing embedded biases will stand out. These rigorously designed datasets pose a formidable challenge, pushing the boundaries of any single approach. Our proposed datasets are challenging that test the limits of any single method across diverse datasets, thereby creating significant opportunities for developing fair graph learning methods and setting new benchmarks in the field. Our main contributions are:

\begin{itemize}[leftmargin=*]
    \item We conduct a thorough examination of existing model selection strategies and widely used datasets within the realm of fair graph learning, identifying key shortcomings that hinder accurate assessment and progress in the field, and then offer our solutions.
    \item We unveil a comprehensive collection of synthetic, semi-synthetic, and real-world datasets specifically crafted for fair graph learning, aiming to cater to diverse needs of research in this domain.
    \item By benchmarking leading fair graph learning approaches on our datasets, we offer valuable insights into their effectiveness, shedding light on the intricacies of achieving fairness in graph-based models.
\end{itemize}

\section{Related Works}

In this section, we review the recent advances on graph neural networks and fair graph learning models.

\subsection{Graph Neural Networks}
Graph neural networks (GNNs) have revolutionized the analysis of graph-structured data across various tasks, including node classification~\cite{kipf2016semi, velickovic2018graph}, graph classification~\cite{sui2022causal}, and link prediction~\cite{zhang2018link,zhao2022learning}. These networks fall into two primary categories: spatial-based GNNs, which utilize direct graph structures to focus on node and neighbor interactions for feature learning, and spectral-based GNNs, which analyze graphs through the spectral domain using the graph Laplacian and its eigenvectors to grasp global graph properties. The exceptional capabilities of GNNs have broadened their application~\cite{hamilton2020graph}, ranging from financial institutions using them to detect fraudulent activities in transaction networks~\cite{daicomprehensive} to their integration into critical decision-making systems where fairness and interpretability become paramount~\cite{yuan2022explainability}. Despite their widespread success, recent research highlights a significant concern: GNNs can exhibit implicit biases towards different groups, potentially leading to unfair outcomes~\cite{dai2021say}. This issue is of particular concern in sensitive applications, underscoring the urgency of incorporating fairness into the GNN modeling process. Bias in GNNs typically arises from two sources: the inherent prejudices present in the input data, and the algorithmic tendencies of GNNs that may favor certain patterns or connections~\cite{dai2023unified}. Consequently, there's a growing movement within the research community towards developing fairer GNN models, aiming to address these biases for more equitable graph-based applications~\cite{ma2022learning}.

\subsection{Fair Graph Learning}
Fairness has become a pivotal issue in machine learning, prominently within the Graph Neural Networks (GNNs) domain~\cite{sekhon2016perceptions, madras2018learning, mehrabi2021survey}. The evaluation of model fairness encompasses various perspectives, such as group fairness~\cite{hardt2016equality}, individual fairness~\cite{dwork2012fairness}, and counterfactual fairness~\cite{kusner2017counterfactual}, all of which are pertinent to GNN assessment~\cite{agarwal2021unified}. In the realm of GNN fairness, concepts like statistical parity~\cite{zemel2013learning} and equal opportunity~\cite{hardt2016equality} are particularly prominent. Efforts such as FairGNN~\cite{dai2021say} employ adversarial training to improve fairness, aiming to prevent the model from leveraging sensitive attributes for predictions. However, the traditional reliance on correlation-based methods for ensuring fairness is challenged by their inability to navigate complexities such as Simpson's paradox. This limitation has spurred a shift towards counterfactual fairness, rooted in causal theory, that promises a deeper, more nuanced approach by focusing on causal relationships to circumvent biases induced by correlations. This shift is exemplified by recent innovations like NIFTY~\cite{agarwal2021unified}, showcasing a keen interest in counterfactual approaches to GNNs. FairVGNN~\cite{wang2022improving} stands out by generating fair views via the automatic identification and masking of sensitive-correlated features, adjusting to correlation changes after feature propagation, thereby achieving state-of-the-art performance on a variety of standard fair graph learning datasets.

\section{Issues of Evaluation Settings}\label{sec: earlystopping}
When evaluating fair graph learning methods, we often care about both the model classification performance and fairness. Specifically, the following metrics are often adopted for evaluation. For evaluating the classification performance, we often utilize metrics such as accuracy (ACC), ROC AUC, and F1-score. To quantify group fairness, we use statistical parity (SP) \cite{dwork2012fairness} and equal opportunity (EO) \cite{hardt2016equality}. The exact metrics are defined as: 
\begin{equation}\label{fairness metric}
\begin{gathered}
\Delta_{S P}=|P(\hat{y}=1 \mid s=0)-P(\hat{y}=1 \mid s=1)|, \\
\Delta_{E O}=|P(\hat{y}=1 \mid y=1, s=0)-P(\hat{y}=1 \mid y=1, s=1)|,
\end{gathered}
\end{equation}
where $y \in \{0,1\}$ denotes the binary label, $s \in \{0,1\}$ denotes the sensitive attribute, and $\hat{y} \in \{0,1\}$ denotes prediction of the classifier. 

\begin{table}[ht]
\centering
\caption{Comparison of different model selection strategies. }
\label{tab: early-stop}
\begin{tabular}{c|c|c|c|c|c}
\toprule
Strategy         & ACC   & AUC   & F1    & $\Delta_{\mathit{SP}}$ & $\Delta_{\mathit{EO}}$ \\ \midrule
Model selection 1 & 69.20 & 62.00 & 81.36 & 2.58                   & 0.63                   \\
Model selection 2 & 69.20 & 64.75 & 80.31 & 8.16                   & 4.62                   \\
Model selection 3 & 61.60 & 64.25 & 68.63 & 5.55                   & 1.02                   \\ \bottomrule
\end{tabular}
\end{table}

Unlike standard node classification models that stop training when their performance on the validation set stops improving significantly, fair graph learning methods need to think about how well they balance accuracy and fairness before deciding when to stop training. To achieve this goal various model selection strategies have been developed:

\begin{itemize}[leftmargin=*]
\item \textbf{Model selection 1}. It first eliminates the training epochs whose ACC and AUC are below the preset thresholds. Then, from the remaining epochs, it picks the ones with the best fairness measures, specifically Statistical Parity (SP) and Equal Opportunity (EO). This strategy is used in the implementation of FairGNN~\cite{dai2021say}.

\item \textbf{Model selection 2}. It uses the validation loss to choose the best model. The model with the lowest validation loss would be tested. This strategy is employed in the NIFTY\cite{agarwal2021unified} implementation.

\item \textbf{Model selection 3}. It uses the validation AUC to select the best model. The model with the highest validation AUC would be tested. NIFTY~\cite{agarwal2021unified} utilized this strategy when replicating FairGNN in its experiment. 
\end{itemize}

\begin{algorithm}
\caption{The Proposed Model Selection Strategy}
\label{alg:early_stop}
\begin{algorithmic}[1]
\State Initialize $best\_fairness$ to $\infty$;
\State Initialize $best\_epoch$ to $0$;
\State Initialize $threshold\_ratios$ to $[0.95, 0.94, 0.93, 0.92, 0.91, 0.9]$;
\For{$ratio$ in $threshold\_ratios$}
    \State $threshold\_acc \gets max\_val\_acc \times ratio$;
    \State $threshold\_roc \gets max\_val\_roc \times ratio$;
    \State $threshold\_f1 \gets max\_val\_f1 \times ratio$;
    
    \For{each $epoch$}
        \If{$val\_acc \geq threshold\_acc$ \textbf{and} $val\_roc \geq threshold\_roc$ \textbf{and} $val\_f1 \geq threshold\_f1$ \textbf{and} $(val\_parity + val\_equality) < best\_fairness$}
            \State $best\_fairness \gets val\_parity + val\_equality$;
            \State $best\_epoch \gets epoch$;
        \EndIf
    \EndFor
\EndFor
\end{algorithmic}
\end{algorithm}

The inconsistency of model selection strategies in the same paper caught our attention. Specifically, in NIFTY~\cite{agarwal2021unified}, NIFTY methods adopt Model selection 2, while FairGNN methods adopt Model selection 3. Table~\ref{tab: early-stop} shows the result of running NIFTY-GCN with these three model selection strategies on the German dataset. Similar phenomena are observed on other datasets and with other models, detailed in Appendix~\ref{appendix: early stop}. These results emphasize an inadequate basis for evaluating these methods fairly caused by the inconsistency of strategies. For evaluating equitable, a unified model selection strategy should be adopted for all methods. 
However, there are some issues with existing strategies. Model selection 2 and 3 do not consider the trade-off between accuracy and fairness. The effectiveness of Model selection 3 is highly dependent on manually preset thresholds, which is certainly cumbersome for evaluating various methods on different datasets. 
Therefore, implementing a consistent and equitable model selection strategy is imperative for the benchmark of fair graph learning methods. 

Our model selection strategy is described in Algorithm~\ref{alg:early_stop}. Compared to the existing strategies, the proposed model selection strategy balances utility and fairness and employs the adaptive thresholds. Since these graph fairness learning methods aim to sacrifice a small portion of utility for higher fairness, the threshold interval is set as $90\% - 95\%$ to trade-off. Additionally, using three classification performance metrics ensures a fair comparison of baseline performance. We anticipate that this standardized model selection approach will assist researchers in achieving a more equitable assessment of fair graph learning models.

\begin{table*}[ht]
\normalsize
\centering
\caption{Statistics of commonly-used Semi-synthetic and Real-world datasets in fair graph learning works.}
\label{original_datasets}
\begin{tabular}{@{}lcccccc@{}}
\toprule
\multicolumn{1}{c}{\textbf{Dataset Type}} & \multicolumn{3}{c}{\textbf{Semi-synthetic}} & \multicolumn{3}{c}{\textbf{Real-world}} \\
\cmidrule(r){2-4} \cmidrule(l){5-7}
\multicolumn{1}{c}{\textbf{Dataset}} & \textbf{German} & \textbf{Bail} & \textbf{Credit} & \textbf{Pokec-z} & \textbf{Pokec-n} & \textbf{NBA} \\ \midrule
\#Nodes & 1,000 & 18,876 & 30,000 & 67,797 & 66,569 & 403 \\
\#Edges & 21,742 & 311,870 & 1,421,858 & 617,958 & 517,047 & 10,621 \\
\#Features & 27 & 18 & 13 & 69 & 69 & 39 \\
Average Degree & 44.48 & 34.04 & 95.79 & 19.23 & 16.53 & 53.71 \\
Sens. & Gender & Race & Age & Region & Region & Nationality \\
Label & Good/bad Credit & Bail/no bail & Payment default/no default & Working Field & Working Field & Salary \\ \bottomrule
\end{tabular}
\end{table*}

\begin{table*}[ht]
    \begin{minipage}{\columnwidth}
        \centering
        \caption{Results of models on Semi-synthetic datasets. $\uparrow$ represents the larger, the better, while $\downarrow$ represents the opposite.}
        \label{tab: old_semi_results}
        \resizebox{\linewidth}{!}{
        \begin{tabular}{lcccc}
        \toprule
        & \textbf{MLP} & \textbf{GCN} & \textbf{FairGNN} & \textbf{NIFTY} \\
        \midrule
        \textbf{Metric} & \multicolumn{4}{c}{\textbf{German}} \\
        \midrule
        ACC ($\uparrow$)       & 71.36 $\pm$ 1.35 & 71.52 $\pm$ 1.17 & 69.84 $\pm$ 0.60 & 70.40 $\pm$ 1.36 \\
        AUC ($\uparrow$)       & 72.45 $\pm$ 0.75 & 73.79 $\pm$ 2.09 & 62.47 $\pm$ 1.56 & 69.3 $\pm$ 1.39 \\
        F1 ($\uparrow$)        & 82.29 $\pm$ 0.25 & 80.01 $\pm$ 1.24 & 82.11 $\pm$ 0.26 & 81.12 $\pm$ 1.91 \\
        $\Delta_{\mathit{SP}}$ ($\downarrow$) & 7.25 $\pm$ 6.35 & 36.67 $\pm$ 11.62 & 1.78 $\pm$ 3.15 & 4.79 $\pm$ 1.68 \\
        $\Delta_{\mathit{EO}}$ ($\downarrow$) & 3.28 $\pm$ 3.16 & 28.78 $\pm$ 9.54 & 2.00 $\pm$ 3.08 & 3.42 $\pm$ 2.45 \\
        \midrule
        \textbf{Metric} & \multicolumn{4}{c}{\textbf{Bail}} \\
        \midrule
        ACC ($\uparrow$)       & 88.13 $\pm$ 0.62 & 84.49 $\pm$ 1.26 & 84.50 $\pm$ 1.07 & 77.68 $\pm$ 7.14 \\
        AUC ($\uparrow$)       & 90.69 $\pm$ 0.74 & 88.76 $\pm$ 1.30 & 89.08 $\pm$ 1.47 & 81.27 $\pm$ 0.86 \\
        F1 ($\uparrow$)        & 82.43 $\pm$ 1.06 & 79.38 $\pm$ 1.29 & 79.50 $\pm$ 1.19 & 69.23 $\pm$ 2.53 \\
        $\Delta_{\mathit{SP}}$ ($\downarrow$) & 0.76 $\pm$ 0.54 & 7.43 $\pm$ 0.91 & 7.32 $\pm$ 0.94 & 5.04 $\pm$ 0.33 \\
        $\Delta_{\mathit{EO}}$ ($\downarrow$) & 4.42 $\pm$ 0.33 & 4.31 $\pm$ 0.96 & 4.25 $\pm$ 0.91 & 4.47 $\pm$ 1.11 \\
        \midrule
        \textbf{Metric} & \multicolumn{4}{c}{\textbf{Credit}} \\
        \midrule
        ACC ($\uparrow$)       & 76.91 $\pm$ 1.94 & 73.58 $\pm$ 0.84 & 73.41 $\pm$ 1.21 & 73.54 $\pm$ 1.84 \\
        AUC ($\uparrow$)       & 71.36 $\pm$ 0.50  & 67.61 $\pm$ 0.27 & 68.99 $\pm$ 0.19 & 68.93 $\pm$ 0.09 \\
        F1 ($\uparrow$)        & 86.32 $\pm$ 2.48 & 82.87 $\pm$ 0.74 & 82.59 $\pm$ 1.06 & 82.67 $\pm$ 1.65 \\
        $\Delta_{\mathit{SP}}$ ($\downarrow$) & 2.26 $\pm$ 4.51 & 11.47 $\pm$ 1.56 & 4.41 $\pm$ 2.71 & 8.56 $\pm$ 0.49 \\
        $\Delta_{\mathit{EO}}$ ($\downarrow$) & 1.78 $\pm$ 3.56 & 9.61 $\pm$ 1.63 & 2.97 $\pm$ 1.99 & 6.44 $\pm$ 0.32 \\
        \bottomrule
        \end{tabular}}
    \end{minipage}\hfill
    \begin{minipage}{\columnwidth}
        \centering
        \caption{Results of models on real-world datasets.$\uparrow$ represents the larger, the better, while $\downarrow$ represents the opposite.}
        \label{tab: old_real_results}
        \resizebox{\linewidth}{!}{
        \begin{tabular}{lcccc}
        \toprule
        & \textbf{MLP} & \textbf{GCN} & \textbf{FairGNN} & \textbf{NIFTY} \\
        \midrule
        \textbf{Metric} & \multicolumn{4}{c}{\textbf{Pokec-z}} \\
        \midrule
        ACC ($\uparrow$)       & 65.18 $\pm$ 1.06 & 69.36 $\pm$ 0.21 & 65.97 $\pm$ 2.09 & 64.47 $\pm$ 1.12 \\
        AUC ($\uparrow$)       & 70.84 $\pm$ 0.79 & 74.86 $\pm$ 0.93 & 70.74 $\pm$ 1.11 & 70.45 $\pm$ 0.66 \\
        F1 ($\uparrow$)        & 65.89 $\pm$ 1.46 & 67.20 $\pm$ 0.51 & 67.13 $\pm$ 0.68 & 65.56 $\pm$ 1.65 \\
        $\Delta_{\mathit{SP}}$ ($\downarrow$) & 2.76 $\pm$ 0.72 & 4.76 $\pm$ 1.00 & 2.41 $\pm$ 1.50 & 3.51 $\pm$ 1.88 \\
        $\Delta_{\mathit{EO}}$ ($\downarrow$) & 1.90 $\pm$ 0.96 & 5.05 $\pm$ 1.11 & 2.15 $\pm$ 1.15 & 2.46 $\pm$ 2.31 \\
        \midrule
        \textbf{Metric} & \multicolumn{4}{c}{\textbf{Pokec-n}} \\
        \midrule
        ACC ($\uparrow$)       & 67.42 $\pm$ 0.36 & 70.15 $\pm$ 0.46 & 65.71 $\pm$ 2.52 & 65.57 $\pm$ 1.31 \\
        AUC ($\uparrow$)       & 72.10 $\pm$ 0.47 & 74.89 $\pm$ 0.19 & 70.40 $\pm$ 2.23 & 68.75 $\pm$ 0.38 \\
        F1 ($\uparrow$)        & 62.23 $\pm$ 1.76 & 65.23 $\pm$ 0.53 & 63.22 $\pm$ 1.60 & 60.21 $\pm$ 1.44 \\
        $\Delta_{\mathit{SP}}$ ($\downarrow$) & 6.57 $\pm$ 1.14 & 7.84 $\pm$ 0.76 & 5.78 $\pm$ 3.14 & 5.66 $\pm$ 0.92 \\
        $\Delta_{\mathit{EO}}$ ($\downarrow$) & 8.67 $\pm$ 0.97 & 11.64 $\pm$ 1.12 & 7.56 $\pm$ 3.27 & 7.28 $\pm$ 1.75 \\
        \midrule
        \textbf{Metric} & \multicolumn{4}{c}{\textbf{NBA}} \\
        \midrule
        ACC ($\uparrow$)       & 67.32 $\pm$ 0.56 & 72.02 $\pm$ 0.70 & 70.33 $\pm$ 0.46 & 62.44 $\pm$ 4.28 \\
        AUC ($\uparrow$)       & 72.48 $\pm$ 0.74 & 76.95 $\pm$ 0.19 & 76.33 $\pm$ 0.45 & 69.27 $\pm$ 1.22 \\
        F1 ($\uparrow$)        & 71.14 $\pm$ 2.31 & 74.41 $\pm$ 1.19 & 74.50 $\pm$ 0.70 & 66.87 $\pm$ 3.51 \\
        $\Delta_{\mathit{SP}}$ ($\downarrow$) & 4.00 $\pm$ 0.92 & 2.03 $\pm$ 0.86 & 1.85 $\pm$ 1.29 & 6.21 $\pm$ 1.88 \\
        $\Delta_{\mathit{EO}}$ ($\downarrow$) & 0.78 $\pm$ 0.44 & 3.32 $\pm$ 1.49 & 1.61 $\pm$ 2.09 & 3.91 $\pm$ 1.89 \\
        \bottomrule
        \end{tabular}}
    \end{minipage}
\end{table*}

\section{Issues of Popular Graph Fairness Datasets}~\label{sec:issues}
Good datasets are essential for advancing the field of study. However, our thorough review reveals that datasets commonly used for fair graph learning suffer from significant issues that could slow progress in this area. To verify these issues, we not only run GNN method and fairness-focused models, but we also include MLP as baseline, which is not always included in existing literature. Moreover, Our experiments set a fine-grained parameter search space for each baseline and uniformly employ our proposed model selection strategy to obtain feasible comparisons.
Our empirical findings, detailed in Tables~\ref{tab: old_semi_results} and \ref{tab: old_real_results}, illustrate these problems. We examine both semi-synthetic and real-world datasets widely used in the community, identifying specific concerns to be addressed.

\subsection{Semi-synthetic Datasets}\label{sec:semi}

We examined semi-synthetic datasets, specifically focusing on the German, Bail, and Credit datasets. The statistical details of these datasets are provided in Table~\ref{original_datasets}. More descriptions of these datasets can be found in Appendix~\ref{sec: old details}. Our experiments led to several concerns detailed below:

\noindent\textbf{Obs 1:} 
Considering predictive capabilities assessed through ACC, AUC and F1, alongside fair metrics such as difference in $\Delta_{\mathit{SP}}$ and $\Delta_{\mathit{EO}}$, graph-based approaches like \textit{GCN do not demonstrate superior performance compared to MLP across various datasets, which may challenge the necessity to use graph-based methods in these datasets.} As shown in Table~\ref{original_datasets}, MLPs achieve comparable predictive accuracy without compromising fairness metrics on three widely used semi-synthetic datasets. Specifically, performance metrics for ACC, AUC, and F1 scores for MLP and GCN are quite close across these datasets. What's worse, MLPs show a significant advantage in terms of $\Delta_{\mathit{SP}}$ and $\Delta_{\mathit{EO}}$, indicating a clear lead in fairness. The result is a strong signal that the graph structures of these semi-synthetic datasets do not contribute meaningful information for enhancing predictions. The rationale behind this is straightforward. According to the dataset generation process described in~\cite{agarwal2021unified}, these datasets originate from tabular data, with graph structures generated based on feature similarity. Thus, reiterating this feature similarity through graph structures does not enrich graph-based methods with novel insights. Moreover, the emphasis on feature similarity in constructing graph structures might inadvertently introduce noise to graph-based models, such as GNNs, potentially deteriorating fairness metrics. These results raise concerns about the necessity of using graph structures for these tasks and suggest that these datasets may not be suitable for fair graph learning problem.

\noindent\textbf{Obs 2}:
\textit{In the analysis of fairness-focused models (FairGNN and NIFTY) versus MLP, we observe no consistent superiority in utility and fairness.} As detailed in Table~\ref{tab: old_semi_results}, a simple MLP model outperforms these fairness-focused models across all evaluated metrics, including ACC, AUC, F1, $\Delta_{\mathit{SP}}$, and $\Delta_{\mathit{EO}}$, by a significant margin. This discrepancy deepens concerns regarding the applicability of datasets for fostering development of fair graph learning algorithms.

\subsection{Real-World Datasets}

Our investigation extends to real-world datasets, with a particular focus on the Pokec-z, Pokec-n, and NBA datasets. The specifics of these datasets are detailed in Table~\ref{original_datasets}, providing a comprehensive statistical overview. More descriptions of these datasets can be seen in Appendix~\ref{sec: old details}. From our experiments, these real-world datasets present several issues:

\noindent \textbf{Obs 1}: As shown in Table \ref{tab: old_real_results}, the bias in topology is not distinctly apparent among the Pokec-z and NBA datasets. The unfairness demonstrated by the GCN model is close to the MLP model. These small values result in the bias-alleviating effect of these fairness methods being limited. In other words, these datasets do not provide sufficient room for comparison of fairness methods. 

\noindent \textbf{Obs 2}: In Pokec-z and Pokec-n datasets, MLP superior fairness methods in classification performance while the unfairness of MLP is similar to, or even less than the fairness methods. This makes one wonder if fair learning methods are needed on these datasets rather than a simple MLP.

\noindent \textbf{Obs 3}: We found that there is a problem with merging multiple labels into binary labels during the processing of real-world datasets. This conversion oversimplifies the inherent complexity of the data, potentially leading to a skewed representation of the original information. Such a approach may result in less robust models, which does not allow for a fair evaluation of the various models.

\subsection{Summary}

Based on the observations, we can conclude that the primary issue across six existing datasets is the lack of meaningful information provided by their graph structures. Consequently, graph-based methods tend to underperform compared to MLP. If graph structure were sufficiently informative, fairness-focused methods would still attain higher accuracy than MLP, albeit model utility is slightly reduced compared to conventional GNN architectures. It is noteworthy that fairness-focused methods also exhibit shortcomings in fairness compared to MLP in certain datasets, highlighting the inadequacy of these datasets for evaluating fair graph learning methods. Hence, future research should consider these limitations when selecting or creating datasets for assessing graph unfairness.

\section{New Fair Graph Learning Datasets}

In light of the issues identified with existing semi-synthetic and real-world datasets, there is a pressing need for new datasets that better benchmark fair graph learning methods. To address these challenges and push the boundaries of fair graph learning research forward, we propose the introduction of new datasets specifically designed to overcome the limitations of current datasets. We seek to offer a more robust and challenging benchmark for developing and evaluating fair graph learning algorithms. This section outlines the development process, characteristics, and potential impact of these new datasets on the field of fair graph learning. Our goal is to facilitate the development of more accurate, fair, and generalizable graph learning models that can navigate the intricacies of real-world social structures and biases.
In the construction of new datasets, we prioritize the following critical considerations:

\begin{table*}[ht]
\centering
\caption{Investigating the Correlation Between Edge Generation Probability and GCN Prediction Accuracy for Different Groups. Symbols "$+$" and "$-$" represent positive and negative correlations, highlighting the variation in accuracy across groups as influenced by edge generation probability. Nodes are classified into four groups based on sensitive attributes and labels ("S0Y0", "S0Y1", "S1Y0", "S1Y1"). Edges are further categorized into ten types according to the characteristics of their connecting nodes (e.g., "S0Y0-S0Y0"), allowing for a detailed analysis of the network's structure and its impact on model performance.}
\label{syn_edge}
\huge
\resizebox{\textwidth}{!}{
\begin{tabular}{c|cccc|cc|cc|cc}
\toprule
 & S0Y0-S0Y0 ($E_1$)  & S0Y1-S0Y1 ($E_2$) & S1Y0-S1Y0 ($E_3$) & S1Y1-S1Y1 ($E_4$) & S0Y0-S1Y0 ($E_5$) & S0Y1-S1Y1 ($E_6$) & S0Y0-S0Y1 ($E_7$) & S1Y0-S0Y1 ($E_8$) & S0Y1-S1Y0 ($E_8$) & S0Y0-S1Y1 ($E_{10}$) \\ \midrule
S0Y0  & \textbf{+} & \textbf{}  & \textbf{}  & \textbf{}  & \textbf{+} & \textbf{}  & \textbf{-} & \textbf{}  & \textbf{}  & \textbf{-} \\
S0Y1  & \textbf{}  & \textbf{+} & \textbf{}  & \textbf{}  & \textbf{}  & \textbf{+} & \textbf{-} & \textbf{}  & \textbf{-} & \textbf{}  \\
S1Y0  & \textbf{}  & \textbf{}  & \textbf{+} & \textbf{}  & \textbf{+} & \textbf{}  & \textbf{}  & \textbf{-} & \textbf{-} & \textbf{}  \\
S1Y1  & \textbf{}  & \textbf{}  & \textbf{}  & \textbf{+} & \textbf{}  & \textbf{+} & \textbf{}  & \textbf{-} & \textbf{}  & \textbf{-} \\ \bottomrule
\end{tabular}}
\end{table*}

\begin{table*}[ht]
\centering
\caption{Results of models for Syn-1 and Syn-2 datasets.}
\label{syn_results}
\resizebox{\textwidth}{!}{
\begin{tabular}{c|c|ccccc|cccc}
\toprule
\multirow{2}{*}{Dataset} & \multirow{2}{*}{Method} & \multirow{2}{*}{ACC ($\uparrow$)} & \multirow{2}{*}{AUC ($\uparrow$)} & \multirow{2}{*}{F1 ($\uparrow$)} & \multirow{2}{*}{$\Delta_{\mathit{SP}}$ ($\downarrow$)} & \multirow{2}{*}{$\Delta_{\mathit{EO}}$ ($\downarrow$)} & \multicolumn{4}{c}{Group   ACC}                           \\
                         &                         &                               &                               &                              &                              &                              & S0Y0         & S0Y1         & S1Y0         & S1Y1         \\ \midrule
\multirow{2}{*}{Syn-1}   & MLP                     & 78.84 ± 0.34                  & 87.25 ± 0.28                  & 80.19 ± 0.40                 & 1.76 ± 1.10                  & 4.35 ± 1.89                  & 81.86 ± 0.74 & 80.63 ± 1.26 & 76.51 ± 1.01 & 76.28 ± 1.24 \\
                         & GCN                     & 86.96 ± 0.66                  & 94.63 ± 0.05                  & 87.90 ± 0.58                 & 10.97 ± 0.85                 & 10.37 ± 1.32                 & 95.40 ± 1.67 & 81.95 ± 1.17 & 78.86 ± 1.28 & 92.33 ± 0.75 \\ \midrule
\multirow{2}{*}{Syn-2}   & MLP                     & 71.04 ± 0.79                  & 78.51 ± 0.66                  & 72.59 ± 1.05                 & 10.52 ± 1.04                 & 7.57 ± 1.31                  & 71.00 ± 3.30 & 73.16 ± 2.18 & 73.55 ± 1.12 & 65.59 ± 2.39 \\
                         & GCN                     & 78.98 ± 0.58                  & 86.69 ± 0.20                  & 80.32 ± 0.57                 & 22.04 ± 1.44                 & 24.09 ± 2.75                 & 77.80 ± 2.91 & 88.50 ± 1.13 & 81.75 ± 1.93 & 64.41 ± 3.04 \\ \bottomrule
\end{tabular}}
\end{table*}

\begin{itemize}[leftmargin=*]
    \item \textbf{Graph Structure Utility.} 
    Graph structure must demonstrably enhance predictive performance, i.e., helpful for prediction task.
    \item \textbf{Bias Amplification through Graph Structure.}
    Graph structure should amplify the bias information. Thus, it can render the performance discrepancy for different fair graph learning methods involving graph structure information.
\end{itemize}
These principles ensure that only models adept at leveraging graph structure for enhanced information processing, while simultaneously mitigating bias inherent within, will excel. Consequently, models relying solely on feature-based methodologies will find themselves at a disadvantage due to their inability to harness the graph structure. Similarly, methods that overlook the bias present in graph structures will face challenges, pushing fairness-oriented models to innovate beyond merely identifying and correcting for bias. This approach aims to foster the development of models that not only capitalize on informational wealth of graph structures but also navigate and neutralize biases effectively, setting a new standard for fairness in graph learning research. Starting from synthetic datasets allows researchers to control utility and bias, and then transition to new semi-synthetic datasets, and finally evaluate models on real-world datasets to provide realistic test scenarios. This progressive benchmarking approach enables a thorough assessment of model capabilities across different stages of dataset realism, ensuring robustness and effectiveness in real-world applications.

\subsection{Synthetic Datasets}
This section explores the relationship between graph structures and fairness performance, outlines the data generation process, and introduces two datasets to demonstrate our analysis framework.

\subsubsection{Interplay Between Edge Generation Probability and Fairness Metrics}\label{sec:5.1.1}
We propose a comprehensive framework, illustrated in Table~\ref{syn_edge}, to explore this interplay. Our focus is on scenarios with binary sensitive attributes and binary labels, where the probability of edge creation directly influences the accuracy of different groups, subsequently impacting fairness metrics. This approach aids in the design and enhancement of synthetic and semi-synthetic datasets. The process unfolds in two pivotal steps:

\begin{itemize}[leftmargin=*]
    \item \textbf{From Edge Generation Probability to Group Accuracy:} 
    The correlation between the probability of generating edges and the accuracy of specific groups is outlined in Table~\ref{syn_edge}. For instance, if we fix the edge generation probabilities for other connections and increase the probability for the "S0Y0-S0Y0" edge, we anticipate an improvement in the accuracy for the "S0Y0" group. This step provides a methodical way to predict group accuracy based on edge generation dynamics.
    
    \item \textbf{From Group Accuracy to Fairness Metrics:}
    The fairness metrics, such as Statistical Parity (SP) and Equal Opportunity (EO), are crucial for assessing fairness. SP gauges the variance in predictive probabilities, whereas EO evaluates the accuracy discrepancy between groups. By examining the accuracy of various groups, we gain insights into potential changes in these fairness metrics, offering a straightforward strategy to assess and enhance fairness in model predictions.
\end{itemize}

\subsubsection{Dataset Construction Process}
This refined approach underscores the significance of understanding the underlying graph structure to inform dataset design and improve fairness in algorithmic decisions. It can also be an effective tool for us to generate synthetic datasets with biased graph structures for fairness problems. We present the generating process of synthetic datasets as follows:

\begin{enumerate}[leftmargin=*]
    \item Generate \(y\) and \(s\) for \(n\) samples from a categorical distribution,
    \begin{align*}
        \qquad (s_i, y_i) \sim \text{Categorical}(p_{00}, p_{01}, p_{10}, p_{11}) \quad \text{for} \quad i=1, \ldots, n,
    \end{align*}
    where \(p_{00}, p_{01}, p_{10}, p_{11}\) denote the probabilities of generating the four possible outcomes for the pairs \((s, y)\), with each pair representing a unique combination of \(s\) and \(y\), ensuring that \(p_{00} + p_{01} + p_{10} + p_{11} = 1\).
    
    \item Generate embeddings $e_y$ and $e_s$ of dimension $d_1$ for $n$ samples from separate multivariate Gaussian distributions,
    \begin{align*}
        e_{y_i=0} &\sim \mathcal{N}\left(-\mu_y, \Sigma_{y}\right) & e_{y_i=1} &\sim \mathcal{N}\left(\mu_y, \Sigma_{y}\right) \\
        e_{s_i=0} &\sim \mathcal{N}\left(-\mu_s, \Sigma_{s}\right) & e_{s_i=1} &\sim \mathcal{N}\left(\mu_s, \Sigma_{s}\right),
    \end{align*}
    where \(\Sigma_{y} = c_1 \cdot I_{d1}\) and \(\Sigma_{s} = c_2 \cdot I_{d1}\) represent the covariance matrices for \(e_y\) and \(e_s\) embeddings, respectively, with $c_1$ and $c_2$ being scalars and \(I_{d1}\) the identity matrix of dimension \(d_1\). The variance (\(\Sigma_{y}\), \(\Sigma_{s}\)) and mean (\(\mu_y\), \(\mu_s\)) parameters are adjustable to modulate the separability between the groups.
    
    \item To construct the node attribute \(x_i\) for each sample, concatenate the embeddings \(e_{y_i}\) and \(e_{s_i}\) as follows:
    \[
        x_i = \left[ e_{y_i} \;|\; e_{s_i} \right],
    \]
    where \(\left[ \cdot \;|\; \cdot \right]\) denotes the concatenation of the \(e_{y_i}\) and \(e_{s_i}\) embeddings, resulting in a single, unified node attribute vector \(x_i\) for each node.

    \item In constructing the graph, we initiate the creation of edges by employing independent Bernoulli distributions for each potential edge type. Specifically, the existence of each edge type \(E_i\) is determined as follows:
    \[
        E_i \sim \text{Bernoulli}(p_i) \quad \text{for each } i = 1, 2, ..., 10,
    \]
    where \(p_i\) represents the probability associated with the formation of edge type \(E_i\). These probabilities correspond to the 10 distinct types of edges enumerated in Table~\ref{syn_edge}, such as "S0Y0-S0Y0", allowing for controlled variability in the graph's connectivity based on predefined probabilities.
    
\end{enumerate}

\begin{table}[ht]
\centering
\caption{The probability of generating different edges in Syn-1 and Syn-2.}
\label{tab: pro of edges}
\begin{tabular}{c|ccccc}
\toprule
\multirow{4}{*}{Syn-1} & $E_1$ & $E_2$ & $E_3$ & $E_4$ & $E_5$    \\ \cline{2-6} 
                       & 0.008 & 0.004 & 0.004 & 0.006 & 0.002    \\ \cline{2-6} 
                       & $E_6$ & $E_7$ & $E_8$ & $E_9$ & $E_{10}$ \\ \cline{2-6} 
                       & 0.002 & 0.002 & 0.002 & 0.001 & 0.002    \\ \midrule
\multirow{4}{*}{Syn-2} & $E_1$ & $E_2$ & $E_3$ & $E_4$ & $E_5$    \\ \cline{2-6} 
                       & 0.006 & 0.008 & 0.007 & 0.005 & 0.002    \\ \cline{2-6} 
                       & $E_6$ & $E_7$ & $E_8$ & $E_9$ & $E_{10}$ \\ \cline{2-6} 
                       & 0.002 & 0.003 & 0.004 & 0.002 & 0.002    \\ \bottomrule
\end{tabular}
\end{table}

\subsubsection{Synthetic Dataset Examples}

We introduce two synthetic datasets, crafted using our framework to illustrate how adjusting parameters such as group ratios, the means and variances of multivariate Gaussian distributions, and edge generation probabilities can influence dataset fairness and performance. These adjustments directly affect the fairness by altering the likelihood of different edge types, as shown in Figure~\ref{syn_edge}, thereby creating disparities in group performance.

Both datasets comprise 5,000 samples, with node attributes dimensioned at 48. Table \ref{tab: pro of edges} displays their graph structure distributions. Analysis of these datasets, detailed in Table \ref{syn_results}, highlights the outcomes of our parameter manipulations.

\noindent\textbf{Syn-1} features a balanced group ratio (\(p_{00}=p_{01}=p_{10}=p_{11}\)). This balance results in minimal performance variance across the four groups when using MLP, enhancing fairness, particularly in terms of Statistical Parity (SP). The introduction of more intra-label group edges not only improves the utility of the graph structure but also enhances GNN performance. Notably, variance in edge types leads to superior GNN outcomes for groups \(s=0,y=0\), and \(s=1,y=1\), albeit introducing some degree of unfairness.

\noindent\textbf{Syn-2} demonstrates an unbalanced group ratio (\(p_{00}=0.22, p_{01}=0.28, p_{10}=0.28, p_{11}=0.22\)), which engenders significant unfairness in MLP predictions. Similar to Syn-1, the graph structure elevates GNN performance. Here, denser connections yield better results for groups \(s=0,y=1\), and \(s=1,y=0\), further contributing to unfairness. By adjusting the Gaussian distribution's variance, we lower MLP's baseline performance, thereby amplifying the graph structure's beneficial impact on performance.

\subsection{Semi-synthetic Datasets}

\begin{table}[ht]
\caption{The proportion of different edges in existing semi-synthetic datasets.}
\label{tab: pro of semi}
\begin{tabular}{c|ccccc}
\toprule
\multirow{4}{*}{German} & $E_1$ & $E_2$ & $E_3$ & $E_4$ & $E_5$    \\ \cline{2-6} 
                        & 0.059 & 0.308 & 0.030 & 0.076 & 0.026    \\ \cline{2-6} 
                        & $E_6$ & $E_7$ & $E_8$ & $E_9$ & $E_{10}$ \\ \cline{2-6} 
                        & 0.087 & 0.246 & 0.085 & 0.036 & 0.047    \\ \midrule
\multirow{4}{*}{Bail}   & $E_1$ & $E_2$ & $E_3$ & $E_4$ & $E_5$    \\ \cline{2-6} 
                        & 0.116 & 0.069 & 0.172 & 0.049 & 0.259    \\ \cline{2-6} 
                        & $E_6$ & $E_7$ & $E_8$ & $E_9$ & $E_{10}$ \\ \cline{2-6} 
                        & 0.110 & 0.058 & 0.058 & 0.048 & 0.061    \\ \midrule
\multirow{4}{*}{Credit} & $E_1$ & $E_2$ & $E_3$ & $E_4$ & $E_5$    \\ \cline{2-6} 
                        & 0.045 & 0.647 & 0.003 & 0.016 & 0.004    \\ \cline{2-6} 
                        & $E_6$ & $E_7$ & $E_8$ & $E_9$ & $E_{10}$ \\ \cline{2-6} 
                        & 0.023 & 0.238 & 0.009 & 0.006 & 0.007    \\ \bottomrule
\end{tabular}
\end{table}

In this section, we statistics the proportion of edges in existing semi-synthetic datasets in Table~\ref{tab: pro of semi}, and then specifically analyze why the graph structure is not sufficiently useful. Then we obtained three new semi-synthetic datasets by adjusting the number of edges following the framework stated in Section~\ref{sec:5.1.1}, and the proportion of edges in new datasets can be found in Appendix~\ref{tab: pro of new semi}.

\begin{table*}[ht]
\centering
\caption{Statistics of new datasets.}
\label{tab: new sta}
\resizebox{\textwidth}{!}{
\begin{tabular}{@{}lccccccc@{}}
\toprule
\multicolumn{1}{c}{\textbf{Dataset}} & \textbf{Syn-1}          & \textbf{Syn-2}          & \textbf{New German}  & \textbf{New Bail}       & \textbf{New Credit}        & \textbf{Sport}      & \textbf{Occupation}   \\ \midrule
\# of nodes                          & 5,000                   & 5,000                   & 1,000                & 18,876                  & 30,000                     & 3,508               & 6,951                 \\
\# of edges                          & 34,363                  & 44,949                  & 20,242               & 31,5870                 & 1,121,858                  & 136,427             & 44,166                \\
\# of features                       & 48                      & 48                      & 27                   & 18                      & 13                         & 768                 & 768                   \\
Sensitive attribute                  & 0/1                     & 0/1                     & Gender (Male/Female) & Race (Black/White)      & Age ($<$25/$>$25)          & Race (White/Black)  & Gender (Male/Female)         \\
Label                                & 0/1                     & 0/1                     & Good/bad Credit      & Bail/no bail            & Payment default/no default & NBA/MLB             & Psy/CS          \\
Average degree                       & 13.75                   & 17.98                   & 41.48                & 34.47                   & 75.79                      & 78.78               & 13.71                 \\
Group Ratio                          & \parbox{2cm}{\centering \begin{tabular}{@{}c@{}}1,218/1,244 \\ 1,239/1,299\end{tabular}} & \parbox{2cm}{\centering \begin{tabular}{@{}c@{}}1,078/1,384 \\ 1,408/1,130\end{tabular}} & \parbox{2cm}{\centering \begin{tabular}{@{}c@{}}191/499 \\ 109/201\end{tabular}}      & \parbox{2cm}{\centering \begin{tabular}{@{}c@{}}5,457/3,860 \\ 6,315/3,244\end{tabular}} & \parbox{2cm}{\centering \begin{tabular}{@{}c@{}}5,906/21,409 \\ 730/1,955\end{tabular}}     & \parbox{2cm}{\centering \begin{tabular}{@{}c@{}}136/1,627 \\ 1,627/118\end{tabular}} & \parbox{2cm}{\centering \begin{tabular}{@{}c@{}}1,751/1,699 \\ 2,951/550\end{tabular}} \\
\# of $E_1 \cup E_2 \cup E_3 \cup E_4$         & 17,225                  & 21,590                  & 12,806               & 170,611                 & 1,013,100                  & 111,736             & 26,138                \\
\# of $E_5\cup E_6$         & 6,319                   & 6,238                   & 2,456                & 75,137                  & 38,592                     & 18,146              & 15,902                \\
\# of $E_7\cup E_8$         & 6,198                   & 10,750                  & 3,192                & 36,210                  & 51,222                     & 1,462               & 1,154                 \\
\# of $E_9\cup E_{10}$         & 4,621                   & 6,371                   & 1,788                & 33,912                  & 18,944                     & 5,083               & 972                   \\ \bottomrule
\end{tabular}
}
\end{table*}

\noindent \textbf{New German Dataset: } As shown in Table 9, the number of intra-label group edges in the German dataset is generally less than the inter-label group edges, except for $E_2$, and the share of $E_7$ is too large. As claimed in Table 5, these problems prevent GNNs from learning effective information through the message-passing mechanism, which leads to lower accuracy. To adjust the proportion of intra- and inter-label group edges and to maintain the difference in accuracy between groups, we randomly reduced 4,000 $E_7$ and randomly added 500 $E_1$, 1,000 $E_3$, and 1,000 $E_4$ to obtain a new German dataset. This new German dataset improves the usefulness of the graph structure while preserving the bias in the graph structure.

\noindent \textbf{New Bail Dataset: } Compared with other datasets, the difference in the proportion of various edges in bail is small, which leads to another problem the features aggregated by GNNs through such an average graph structure would be not discriminative. So it appears that the classification performance of GCN is much worse than that of MLP in Table~\ref{tab: old_semi_results}. For GNNs to better aggregate features, we randomly reduced 40,000 $E_5$ and randomly added 15,000 $E_2$, 20,000 $E_3$, and 4,000 $E_4$. We expect GNNs would perform better on the new bail dataset. 

\noindent \textbf{New Credit Dataset: } Similar to the case of the German dataset, a large proportion of $E_7$ in the Credit dataset decreases the performance of GNNs. We randomly reduced 30,000 $E_7$ to construct a graph structure with a more reasonable proportion of intra- and inter-label group edges. The following GNN classification results demonstrate the effectiveness of these simple adjustments.

\subsection{Real-world Datasets from Twitter}
We have constructed two novel datasets by leveraging the Twitter API, offering insights into real-world social dynamics and biases. These datasets, detailed below, serve as the foundation for our studies on bias mitigation and the robustness of predictive models.

\noindent\textbf{Sport Dataset:} Derived from Twitter, this dataset focuses on athletes in the NBA and MLB. We mapped players to their Twitter accounts, using these accounts as nodes. Edges represent following relationships between players. The sensitive attribute under consideration is the players' race, categorized as either black or white. The objective is to predict the sport of a player (NBA or MLB) without bias influenced by racial attributes. For node features, we aggregated the first five tweets from each player's account and utilized average of their BERT embeddings~\cite{devlin2018bert} as feature representations.

\noindent\textbf{Occupation Dataset:} This dataset also originates from Twitter, with nodes representing users and edges indicating follow relationships. The focus is on users identified within the fields of computer science or psychology. User selection was stratified across multiple layers: starting from a randomly chosen set of users (1st layer), we expanded the dataset by including their followers (2nd layer) and repeated this process up to six layers to ensure diversity. The sensitive attribute here is gender, with the aim to predict a user's field of work without gender-based bias. Node features were derived similarly to the Sport dataset, using the mean of BERT embeddings from the users' tweets.

\section{Benchmarking on New Datasets}

\begin{table*}[ht]
\centering
\caption{Results of models for new datasets.}
\label{tab: new_dataset_results}
\begin{tabular}{c|l|ccccc}
\toprule
Dataset                     & \multicolumn{1}{c|}{Method} & ACC($\uparrow$)       & AUC($\uparrow$)       & F1($\uparrow$)        & $\Delta_{\mathit{SP}}$ ($\downarrow$) & $\Delta_{\mathit{EO}}$ ($\downarrow$) \\ \midrule
\multirow{4}{*}{Syn-1}      & MLP                         & 78.84 ± 0.34 & 87.25 ± 0.28 & 80.19 ± 0.40 & 1.76 ± 1.10                           & 4.35 ± 1.89                           \\
                            & GCN                         & 86.96 ± 0.66 & 94.63 ± 0.05 & 87.90 ± 0.58 & 10.97 ± 0.85                          & 10.37 ± 1.32                          \\ \cdashline{3-7}
                            & FairGNN                     & 85.06 ± 0.42 & 93.07 ± 0.22 & 85.87 ± 0.45 & 1.77 ± 1.03                           & 2.92 ± 1.98                           \\
                            & NIFTY                       & 80.22 ± 2.23 & 88.52 ± 2.23 & 81.45 ± 2.23 & 15.77 ± 6.1                           & 15.65 ± 7.24                          \\ \midrule
\multirow{4}{*}{Syn-2}      & MLP                         & 71.04 ± 0.79 & 78.51 ± 0.66 & 72.59 ± 1.05 & 10.52 ± 1.04                          & 7.57 ± 1.31                           \\
                            & GCN                         & 78.98 ± 0.58 & 86.69 ± 0.20 & 80.32 ± 0.57 & 22.04 ± 1.44                          & 24.09 ± 2.75                          \\ \cdashline{3-7}
                            & FairGNN                     & 74.74 ± 1.04 & 82.82 ± 0.64 & 77.19 ± 1.02 & 1.17 ± 0.43                           & 1.79 ± 1.25                           \\
                            & NIFTY                       & 73.06 ± 0.72 & 80.10 ± 0.33 & 74.46 ± 0.93 & 31.87 ± 3.84                          & 31.47 ± 3.76                          \\ \midrule
\multirow{4}{*}{New German} & MLP                         & 71.36 ± 1.35 & 72.45 ± 0.75 & 82.29 ± 0.25 & 7.25 ± 6.35                           & 3.28 ± 3.16                           \\
                            & GCN                         & 82.08 ± 1.55 & 87.19 ± 1.04 & 88.23 ± 1.01 & 24.52 ± 3.50                          & 6.20 ± 4.20                           \\ \cdashline{3-7}
                            & FairGNN                     & 77.52 ± 3.98 & 83.81 ± 3.54 & 85.34 ± 2.40 & 18.85 ± 7.69                          & 6.68 ± 3.63                           \\
                            & NIFTY                       & 74.4 ± 4.40  & 80.13 ± 3.36 & 83.50 ± 1.95 & 4.33 ± 3.01                           & 2.21 ± 1.74                           \\ \midrule
\multirow{4}{*}{New Bail}   & MLP                         & 88.13 ± 0.62 & 90.69 ± 0.74 & 82.43 ± 1.06 & 0.76 ± 0.54                           & 4.42 ± 0.33                           \\
                            & GCN                         & 92.21 ± 1.33 & 96.37 ± 2.47 & 89.84 ± 2.14 & 9.99 ± 1.48                           & 4.45 ± 2.11                           \\ \cdashline{3-7}
                            & FairGNN                     & 91.61 ± 1.52 & 95.78 ± 1.16 & 89.06 ± 1.86 & 9.25 ± 0.65                           & 4.99 ± 1.35                           \\
                            & NIFTY                       & 80.51 ± 5.85 & 87.02 ± 1.43 & 75.45 ± 3.41 & 5.4 ± 2.22                            & 3.67 ± 0.96                           \\ \midrule
\multirow{4}{*}{New Credit} & MLP                         & 76.91 ± 1.94 & 71.36 ± 0.50 & 86.32 ± 2.48 & 2.26 ± 4.51                           & 1.78 ± 3.56                           \\
                            & GCN                         & 82.61 ± 1.06 & 91.52 ± 1.67 & 87.80 ± 0.74 & 16.86 ± 1.38                          & 24.27 ± 1.54                          \\ \cdashline{3-7}
                            & FairGNN                     & 79.02 ± 0.94 & 84.27 ± 2.69 & 86.98 ± 0.56 & 6.76 ± 6.93                           & 8.28 ± 8.58                           \\
                            & NIFTY                       & 75.73 ± 2.37 & 76.41 ± 1.18 & 84.11 ± 2.58 & 6.85 ± 4.09                           & 5.37 ± 3.89                           \\ \midrule
\multirow{4}{*}{Sport}      & MLP                         & 66.92 ± 1.64 & 73.46 ± 1.80 & 66.87 ± 1.69 & 30.44 ± 3.43                          & 9.00 ± 3.17                           \\
                            & GCN                         & 95.16 ± 0.73 & 98.67 ± 0.35 & 95.22 ± 0.70 & 81.13 ± 1.10                          & 3.46 ± 1.17                           \\ \cdashline{3-7}
                            & FairGNN                     & 94.53 ± 0.73 & 98.71 ± 0.41 & 94.41 ± 0.83 & 78.49 ± 1.80                          & 2.21 ± 1.91                           \\
                            & NIFTY                       & 88.95 ± 4.69 & 96.84 ± 0.45 & 89.59 ± 3.87 & 70.21 ± 7.14                          & 4.12 ± 1.60                           \\ \midrule
\multirow{4}{*}{Occupation} & MLP                         & 78.59 ± 0.55 & 85.18 ± 0.43 & 61.90 ± 1.97 & 21.43 ± 1.61                          & 13.08 ± 2.64                          \\
                            & GCN                         & 81.70 ± 0.56 & 87.89 ± 0.47 & 69.96 ± 1.72 & 25.24 ± 0.95                          & 16.04 ± 1.76                          \\ \cdashline{3-7}
                            & FairGNN                     & 80.92 ± 0.42 & 86.47 ± 0.19 & 67.21 ± 0.50 & 22.75 ± 0.65                          & 14.88 ± 2.04                          \\
                            & NIFTY                       & 78.09 ± 1.11 & 83.27 ± 0.59 & 60.25 ± 4.88 & 20.89 ± 2.36                          & 17.52 ± 1.53                          \\ \bottomrule
\end{tabular}
\end{table*}

This section outlines our empirical investigation designed to assess the utility and integrity of the newly developed datasets. The statistics details of these new datasets are shown in Table~\ref{tab: new sta}. Our goal is to scrutinize the datasets through a series of experiments aimed at addressing the following pivotal questions:
\begin{itemize}[leftmargin=*]
    \item \textbf{(RQ 1)} 
    Are the proposed datasets capable of yielding significant insights and enhancing predictive performance within their graph structures?
    \item \textbf{(RQ 2)} 
    Does the graph structure exhibit biased information, necessitating a proficient model that can adeptly harness the graph's structure while also mitigating any inherent biases?
    \item \textbf{(RQ 3)} 
    Can we gain insights into the commonly used methods with our datasets?
\end{itemize}

This section delineates our experimental evaluation, conducted to ascertain the efficacy of our newly introduced datasets in facilitating fair graph benchmarking. Our experiments are designed to benchmark existing models, thereby providing insights into their performance when applied to diverse and challenging scenarios.

\subsection{Experimental Setup}

In our benchmarking, we selected key fair graph learning methods, including FairGNN~\cite{dai2021say}, which uses an adversarial method with a sensitive feature estimator for fairness, and NIFTY~\cite{agarwal2021unified}, employing a novel augmentation for counterfactual fairness through contrastive learning. Both methods are based on GCN~\cite{kipf2016semi} to leverage graph structure. We also compared these with a standard GCN to understand the role of graph topology and an MLP to gauge the benefit of incorporating graph structure.

To ensure a fair and comprehensive comparison, we meticulously fine-tuned the hyperparameters for each model, tailored to their optimal performance on our datasets. The specifics regarding these configurations are provided in Appendix~\ref{appendix: hyperparameter}. Aligning with the setup in NIFTY~\cite{agarwal2021unified}, our experiments utilize a one-layer GCN for encoding, complemented by a linear layer that functions as both classifier and discriminator. The data partition strategy is consistent with established protocols~\cite{agarwal2021unified, dai2021say}, ensuring comparability. To account for variability in initialization, we report the average results over five runs, each with a unique random seed.

\subsection{Evaluation and Results}

Table~\ref{tab: new_dataset_results} presents the performance outcomes of various methods across the datasets we introduce. These datasets are specifically designed to test the capabilities of fair graph learning approaches, revealing several key attributes conducive to their evaluation. For \textbf{RQ 1}, a comparison between MLP and GCN demonstrates significant enhancements in predictive accuracy, as evidenced by improvements in ACC, AUC, and F1 score across all seven datasets. This distinction is particularly notable when juxtaposed with the outcomes from semi-synthetic and real-world datasets discussed in Section~\ref{sec:issues}, underscoring the value of incorporating graph structures into the analysis for augmenting model performance.  

For \textbf{RQ 2}, an analysis of MLP and GCN's performance based on fairness metrics $\Delta_{\mathit{SP}}$ and $\Delta_{\mathit{EO}}$ reveals that incorporating graph structures often leads to a significant reduction in fairness across all seven datasets. This observation underscores the inherent trade-off in using graph data: while it can enhance model performance, it also risks compromising fairness. Nonetheless, a comparative assessment of fairness-focused algorithms demonstrates notable improvements in fairness metrics for most datasets. Specifically, FairGNN shows comprehensive enhancements across all indicators—ACC, AUC, F1 score, $\Delta_{\mathit{SP}}$, and $\Delta_{\mathit{EO}}$—on Syn-1 and Syn-2. Similarly, NIFTY exhibits parallel improvements on the New German dataset. This trend suggests the feasibility of leveraging graph structures to boost predictive accuracy while simultaneously mitigating bias. Such capability is crucial for fairness benchmarking datasets, serving as a critical measure to evaluate a model's ability to exploit graph data beneficially without sacrificing fairness. These findings indicate that with carefully designed fair graph learning approaches, it is possible to balance both predictive performance and fairness objectives effectively.

In the context of \textbf{RQ 3}, it is apparent that current fairness methodologies struggle to consistently excel across all datasets. This challenge sets a new benchmark, urging further innovation in model development. For example, in the New Bail dataset, FairGNN leverages graph structures to surpass MLP in terms of ACC, AUC, and F1 scores, yet it must navigate the inherent biases within the graph data, leading to a decrease in $\Delta_{\mathit{SP}}$ and $\Delta_{\mathit{EO}}$. While FairGNN manages to mitigate bias more effectively than the baseline GCN, this adjustment comes at the cost of reduced predictive accuracy. Conversely, NIFTY faces difficulties in optimally exploiting graph information, resulting in performance deficits across all metrics, even trailing behind MLP. Our evaluation presents a set of challenging tasks, making it difficult for any single method to excel across all datasets. This situation offers substantial opportunities for the development of strong fair graph learning methods, paving the way for new milestones in the field.

\section{Conclusion}

In conclusion, our exploration into fair graph learning underscores the critical importance of representative datasets for evaluating the performance of fair graph learning methods. Through this work, we have identified a significant gap in the quality and applicability of existing semi-synthetic and real-world datasets. Our findings reveal that, in many cases, simple models such as MLPs can surpass more complex GNNs when the datasets lack meaningful graph structures. To address these shortcomings, we have developed a comprehensive suite of synthetic, semi-synthetic, and real-world datasets designed with the explicit aim of facilitating a fair and rigorous evaluation of fair graph learning methods. These datasets are carefully crafted to encompass critical graph structures and bias information, challenging models to not only leverage graph structures for enhanced prediction accuracy but also to effectively address and mitigate bias inherent in the data. We introduce a unified framework for analyzing the edge generation probability to fairness metrics. Based on this, we provide controllable bias parameters in synthetic and semi-synthetic datasets, allowing researchers to tailor the datasets to specific research needs and bias considerations. Our systematic evaluation of these newly proposed datasets has yielded extensive experimental insights. This work lays the foundation for future progress in fair graph learning, promoting the creation of models that effectively harness graph structures while prioritizing fairness. We introduce challenging tasks that test the limits of any single method across diverse datasets, thereby creating significant opportunities for developing fair graph learning methods and setting new benchmarks in the field.

\section{Acknowledgements}
This material is based upon work supported by the National Science
Foundation (NSF) Grant \#2406648 and \#2406647, Army Research Office (ARO) under grant number W911NF-21-1- 0198, Department of Homeland Security (DHS) CINA under grant number E205949D, and Cisco Faculty Research Award.

\bibliographystyle{ACM-Reference-Format}
\balance
\bibliography{acmart}

\newpage
\appendix
\section{Detailed Issues}
\subsection{Different Model Selection Strategy}\label{appendix: early stop}

We compare the different model selection strategies under consistent conditions among several datasets. As shown in Table~\ref{appendix: tab:earlystop1} and Table~\ref{appendix: tab:earlystop2}, the performance gap caused by the different strategies illustrates the importance of standardizing the model selection strategy.

\begin{table}[ht]
\centering
\caption{Compare the performance of FairGNN using different strategies on real-world datasets.}
\label{appendix: tab:earlystop1}
\resizebox{\linewidth}{!}{
\begin{tabular}{c|c|ccccc}
\toprule
Dataset          & Strategy & ACC & AUC & F1 & $\Delta_{\mathit{SP}}$ & $\Delta_{\mathit{EO}}$ \\ \midrule
\multirow{4}{*}{Pokec\_z} & Strategy 1        & 66.76        & 73.11        & 62.44       & 1.15        & 1.51        \\
                          & Strategy 2        & 68.59        & 73.85        & 67.97       & 1.02        & 2.54        \\
                          & Strategy 3        & 68.71        & 73.91        & 68.36       & 1.34        & 3.29        \\
                          & Ours              & 68.24        & 73.47        & 68.15       & 0.84        & 0.91        \\ \midrule
\multirow{4}{*}{Pokec\_n} & Strategy 1        & 68.91        & 73.25        & 65.56       & 8.08        & 9.99        \\
                          & Strategy 2        & 68.35        & 73.24        & 65.70       & 8.11        & 10.42       \\
                          & Strategy 3        & 68.32        & 73.37        & 66.51       & 8.98        & 9.87        \\
                          & Ours              & 68.91        & 73.25        & 65.56       & 8.08        & 9.99        \\ \midrule
\multirow{4}{*}{NBA}      & Strategy 1        & 69.01        & 76.73        & 73.60       & 1.78        & 0.99        \\
                          & Strategy 2        & 69.01        & 78.16        & 68.87       & 6.98        & 16.31       \\
                          & Strategy 3        & 47.89        & 79.80        & 0.00        & 0.00        & 0.00        \\
                          & Ours              & 70.89        & 76.93        & 74.80       & 1.42        & 0.99        \\ \bottomrule
\end{tabular}}
\end{table}

\begin{table}[ht]
\centering
\caption{Compare the performance of NIFTY using different strategies on semi-synthetic datasets.}
\label{appendix: tab:earlystop2}
\resizebox{\linewidth}{!}{
\begin{tabular}{c|c|ccccc}
\toprule
Dataset          & Strategy & ACC & AUC & F1 & $\Delta_{\mathit{SP}}$ & $\Delta_{\mathit{EO}}$ \\ \midrule
\multirow{4}{*}{German} & Strategy 1        & 70.00        & 65.49        & 81.84       & 1.67        & 0.21        \\
                        & Strategy 2        & 71.20        & 67.86        & 82.44       & 1.09        & 2.42        \\
                        & Strategy 3        & 65.60        & 68.50        & 73.46       & 6.86        & 0.21        \\
                        & Ours              & 72.00        & 70.32        & 83.09       & 1.47        & 0.11        \\ \midrule
\multirow{4}{*}{Bail}   & Strategy 1        & 81.10        & 81.34        & 68.46       & 4.85        & 4.41        \\
                        & Strategy 2        & 80.12        & 79.73        & 67.32       & 4.77        & 4.02        \\
                        & Strategy 3        & 70.31        & 83.37        & 67.79       & 2.06        & 2.52        \\
                        & Ours              & 81.65        & 81.81        & 71.00       & 4.88        & 4.05        \\ \midrule
\multirow{4}{*}{Credit} & Strategy 1        & 68.51        & 69.15        & 78.01       & 10.08       & 9.30        \\
                        & Strategy 2        & 74.20        & 69.19        & 83.24       & 8.76        & 6.68        \\
                        & Strategy 3        & 60.35        & 69.28        & 69.43       & 13.31       & 13.47       \\
                        & Ours              & 74.67        & 69.29        & 83.64       & 8.72        & 6.17        \\ \bottomrule
\end{tabular}}
\end{table}

\subsection{Dataset Details}
\subsubsection{Detailed Existing Datasets}\label{sec: old details}
Here we present a detailed description of six wildly-used datasets we used to validate our proposed issues as follows:
\begin{itemize}
    \item \textbf{German Credit} (German): This dataset models clients as nodes, where edges reflect a high similarity in credit account activities. The objective is to classify individuals into high or low-credit risk categories, considering gender as the sensitive attribute.
    
    \item \textbf{Recidivism} (Bail): It comprises nodes representing defendants who were released on bail between 1990 and 2009. Edges are drawn between nodes with similar criminal records and demographic characteristics. The classification challenge involves predicting bail status based on the sensitive attribute of race.
    
    \item \textbf{Credit Defaulter} (Credit): In this dataset, nodes symbolize credit card users, connected by edges that indicate similarity in purchasing and payment behaviors. The classification goal is to identify users likely to default on payments, with age serving as the sensitive attribute.
    
    \item \textbf{Pokec:} A widely recognized dataset from the Slovak social network, anonymized in 2012, segmented into two subsets: Pokec-z and Pokec-n. These subsets represent user profiles from two significant regions within Slovakia, designated by their respective provinces. The datasets use the geographical region of the users as the sensitive attribute, aiming to predict the employment sector of the users.
    
    \item \textbf{NBA:} Comprising data on roughly 400 NBA players, this dataset uses a player's nationality (categorized into U.S. or non-U.S.) as the sensitive attribute. The dataset constructs a social graph of NBA players through their interactions on Twitter, with the predictive task focusing on determining if a player's salary is above or below the league median.
\end{itemize}

\subsubsection{Detailed New datasets}\label{sec: new details}

The proportion of different edges in new semi-synthetic dataset is shown in Table~\ref{tab: pro of new semi}.

\begin{table}[ht]
\caption{The proportion of different edges in new semi-synthetic datasets.}
\label{tab: pro of new semi}
\begin{tabular}{@{}c|ccccc@{}}
\toprule
\multirow{4}{*}{New German} & $E_1$ & $E_2$ & $E_3$ & $E_4$ & $E_5$    \\ \cline{2-6} 
                            & 0.088 & 0.331 & 0.082 & 0.131 & 0.028    \\ \cline{2-6}  
                            & $E_6$ & $E_7$ & $E_8$ & $E_9$ & $E_{10}$ \\ \cline{2-6} 
                            & 0.093 & 0.067 & 0.091 & 0.038 & 0.050    \\ \cline{1-6}
\multirow{4}{*}{New Bail}   & $E_1$ & $E_2$ & $E_3$ & $E_4$ & $E_5$    \\ \cline{2-6} 
                            & 0.115 & 0.115 & 0.233 & 0.077 & 0.129    \\ \cline{2-6} 
                            & $E_6$ & $E_7$ & $E_8$ & $E_9$ & $E_{10}$ \\ \cline{2-6} 
                            & 0.108 & 0.057 & 0.058 & 0.047 & 0.060    \\ \cline{1-6}
\multirow{4}{*}{New Credit} & $E_1$ & $E_2$ & $E_3$ & $E_4$ & $E_5$    \\ \cline{2-6} 
                            & 0.057 & 0.822 & 0.004 & 0.021 & 0.006    \\ \cline{2-6} 
                            & $E_6$ & $E_7$ & $E_8$ & $E_9$ & $E_{10}$ \\ \cline{2-6}  
                            & 0.029 & 0.034 & 0.012 & 0.008 & 0.009    \\ \bottomrule
\end{tabular}
\end{table}

\section{Experimental Settings}

\subsection{Hyperparameter Selection}\label{appendix: hyperparameter}
Since different methods have different model architectures, their hyperparameters are various and are described respectively as follows:

\noindent \textbf{MLP}: the number of layers $\{2, 3, 4, 5\}$, the number of hidden unit 16, learning rate $\{1e-2, 1e-3, 1e-4\}$, weight decay $\{1e-4, 1e-5\}$, dropout $\{0, 0.5, 0.8\}$.

\noindent \textbf{GCN}: the number of layers $\{1, 2, 3\}$, the number of hidden unit 16, learning rate $\{1e-2, 1e-3, 1e-4\}$, weight decay $\{1e-4, 1e-5\}$, dropout $\{0, 0.5, 0.8\}$.

\noindent \textbf{FairGNN}: the number of hidden unit 32, learning rate $\{1e-2, 1e-3, 1e-4\}$, weight decay $\{1e-4, 1e-5\}$, dropout $\{0, 0.5, 0.8\}$, regularization coefficients $\alpha \{4, 5, 50, 100\}$ and $\beta \{0.01, 1, 5, 20\}$.

\noindent \textbf{NIFTY}: the number of hidden unit 16, project hidden unit 16, drop edge rate 0.001, drop feature rate 0.1, learning rate $\{1e-2, 1e-3, 1e-4\}$, weight decay $\{1e-4, 1e-5\}$, dropout $\{0, 0.5, 0.8\}$, regularization coefficient $\{0.2, 0.4, 0.6, 0.8\}$.

\end{document}